\documentclass[letterpaper, 10 pt, conference]{ieeeconf}
\IEEEoverridecommandlockouts
\usepackage{algorithm}
\usepackage{algpseudocode}
\usepackage{graphicx}
\usepackage{amsmath,amssymb}
\usepackage{booktabs}
\usepackage{array}
\usepackage{siunitx}
\usepackage{subcaption}
\newcommand{\tablestyle}{%
\centering\small
  \renewcommand{\arraystretch}{1.18}%
  \setlength{\extrarowheight}{1pt}%
  \setlength{\tabcolsep}{4pt}%
}
\usepackage[nocheckfootnote]{flushend}
\usepackage{placeins}
\usepackage{tikz}
\usepackage[hidelinks]{hyperref}
\usepackage{microtype}
\usepackage{etoolbox}
\patchcmd{\abstract}{---\,}{:\ }{}{}
\usetikzlibrary{arrows.meta,positioning,shapes.geometric,calc,fit,backgrounds}

\title{\LARGE\bf RoboRacer Arena: Specification-Driven Track\\Construction for Autonomous Racing}

\author{%
    Mihaela-Larisa Clement$^{1,2}$,
    Agnes Poks$^{1}$, and
    Ezio Bartocci$^{1}$%
    \thanks{$^{1}$TU Wien, Vienna, Austria.}%
    \thanks{$^{2}$AIT Austrian Institute of Technology,
    Vienna, Austria.}%
}

\begin{document}

\maketitle
\thispagestyle{empty}
\pagestyle{empty}
\raggedbottom

\begin{abstract}
Learning-based autonomous racing relies on diverse training environments, yet constructing a new track requires geometric design, validation, and simulation-asset generation. This coupling makes track geometry difficult to vary systematically during policy training and evaluation. To address this limitation, we present RoboRacer Arena, a specification-driven pipeline that exposes track geometry as an explicit experimental variable. Starting from natural-language requirements, a seeded coverage-guided constructor generates closed-loop layouts, validates the exported occupancy maps, and automatically builds the corresponding Isaac Sim environments. The same interface admits recorded maps and scaled circuits, yielding an initial reference collection of 130 tracks. Across our evaluation, RoboRacer Arena achieves the highest valid-map generation rate among the tested construction procedures under their respective computational budgets and converts eight benchmark maps into simulation assets in less than 2.5 seconds each. We further reconstruct the RoboRacer vehicle as a CAD and USD asset and use it for parallel residual-policy training and deployment on the physical platform. In physical experiments, policies trained with RoboRacer Arena complete ten consecutive laps at command settings up to four times the nominal training speed. These results demonstrate that explicit track requirements can be connected to validated simulation assets and physical evaluation within a reproducible autonomous-racing workflow.\end{abstract}

\section{Introduction}

\begin{figure*}[!t]
\centering
\resizebox{\textwidth}{!}{\input{figures/fig_pipeline_revised.tex}}
\caption{Overview of RoboRacer Arena, from track specification and map import to parallel policy training and physical deployment. The vehicle asset is reconstructed from component scans and shared across the simulation and physical workflows.}
\label{fig:pipeline}
\end{figure*}

RoboRacer, formerly F1TENTH, provides autonomous-racing researchers with a shared 1:10-scale vehicle platform~\cite{okelly2020f1tenth}. Learning-based control, however, also depends on access to varied training environments~\cite{cobbe2020procgen}. Preparing each racing track currently couples layout design, geometric validation, and simulation-asset construction. A requested layout may need to fit a particular hall, contain prescribed corner types, and maintain a given width and lap length. Repeating these steps for every combination of requirements prevents track geometry from being varied as readily as other experimental parameters.

To address this limitation, we automate the construction process while retaining explicit control over the geometry that reaches the simulator. First, a typed specification represents track dimensions, corner features, and feature order. A seeded geometric search then constructs candidate loops and exports occupancy maps that distinguish free track, occupied walls, and unknown background. Crucially, acceptance is determined on the exported map rather than on the continuous centreline alone, thereby accounting for geometric changes introduced by rasterisation before USD construction.

RoboRacer Arena combines these stages in the workflow depicted in Fig.~\ref{fig:pipeline}. At the core of our approach is a direct connection between specified requirements, validation of exported geometry, and automatic construction of simulation assets. A common occupancy-map interface also admits community maps and imported circuits, which we combine with a CAD-modelled vehicle for parallel learning. Using an established residual-policy architecture, we evaluate the complete workflow from requirement-guided construction to deployment on the corresponding physical car.

The contributions of this work are fourfold:
\begin{enumerate}
\item \textbf{Track construction from explicit requirements.} We introduce a typed specification that guides seeded construction and feature insertion, while geometric checks on the exported occupancy map determine acceptance (Sec.~\ref{sec:req}).
\item \textbf{A common route from maps to simulation.} We develop a map-to-USD builder that constructs separate visual and collision geometry for generated and imported maps, supporting a reference collection of 130 tracks (Sec.~\ref{sec:system}).
\item \textbf{Evaluation across the complete workflow.} We measure specification compliance, asset-construction cost, and training throughput, and subsequently examine tracking and lap completion after physical policy deployment (Sec.~\ref{sec:evaluation}).
\item \textbf{Resources for reproducible experiments.} Upon acceptance, we will release the Isaac Sim extension, RoboRacer CAD and USD vehicle models, construction and validation tools, vehicle configurations, and track collection with documented sources, enabling others to reproduce and extend the workflow.
\end{enumerate}

\section{Related Work}

Procedural generation has been used to construct racing tracks through evolutionary search~\cite{loiacono2011autotrack} and to vary layouts for domain randomisation~\cite{behrens2020trackgen}. Language-driven systems extend environment generation to 3D scenes in Holodeck~\cite{yang2024holodeck} and to simulation and learning tasks in RoboGen~\cite{wang2024robogen}. In contrast, our work targets racing requirements that can be evaluated directly on an exported occupancy map, including dimensions, topology, and corner features. We build on spanning-tree coverage~\cite{gabriely2001spanning}, while feature insertion, circular-arc smoothing, and raster-level validation adapt this construction principle to closed racing corridors.

Occupancy grids provide an established representation for map-based environment construction~\cite{elfes1989occupancy}, while Map2gazebo converts occupied cells into extruded meshes~\cite{map2gazebo}. We use the same class of input to construct racing environments in Isaac Sim, but separate visual barriers from collision boundaries before assembling the floor and materials. This shared representation connects geometric acceptance to asset construction and, at the same time, allows recorded maps to enter the same simulation workflow.

Existing simulators support complementary parts of the racing workflow. F1TENTH Gym provides planar map-based simulation~\cite{okelly2020f1tenth}, whereas AutoDRIVE includes 3D dynamics and rendering~\cite{samak2023autodrive}. Isaac Lab enables parallel GPU simulation~\cite{mittal2025isaaclab}, including applications to small wheeled robots~\cite{han2025wheeledlab}. RoboRacer Arena contributes specification-driven track construction and map-to-asset generation to this setting, while our evaluation adopts the residual Pure-Pursuit architecture of~\cite{trumpp2023rpl}. Raceline-informed learning~\cite{evans2023tal} and physical policy transfer~\cite{brunnbauer2022latent} address complementary questions of controller design and hardware transfer.

\section{Methodology}
\label{sec:req}

Our method connects a typed description of the requested track, a seeded search for candidate geometry, and validation of the exported occupancy map. Since accepted maps pass directly to the USD builder (Sec.~\ref{sec:map2env}), acceptance is determined on the same representation used to construct the simulation environment.

\subsection{Track Specification and Feasibility}

We first express each request through a 12-field schema comprising the track name, lap length $L$, track width $w$, hall length $B_x$, hall width $B_y$, driving direction, hairpin count, chicane count, sweeper count, feature order, minimum corner radius, and maximum straight length. A local LLM instantiates this schema from natural language using schema-constrained decoding at zero temperature. Generation seeds and rasterisation parameters remain independent of this extraction, preventing language-model variation from altering the geometric search itself.

The typed schema allows geometric requirements to be checked independently of the model that extracted them. Unspecified fields receive fixed defaults; for example, the minimum centreline radius is $r_{\min}=\max(\SI{0.9}{\metre},0.75w)$ and the maximum straight length is $\min(B_x,B_y)$. Before geometric search begins, we screen the completed specification using bounds on corridor packing and on the lap length consumed by the requested features.

To estimate the lap length that can fit within the hall, we define an effective corridor pitch $p$ and usable area $U$. Given wall thickness $t_{\mathrm{w}}$, corridor clearance $d_{\mathrm{c}}$, and hall margin $m_{\mathrm{h}}$, the resulting capacity estimate is $L_{\max}=U/p$:
\begin{equation}
\begin{aligned}
p &= w+2t_{\mathrm{w}}+d_{\mathrm{c}},\\
U &= (B_x-2m_{\mathrm{h}})(B_y-2m_{\mathrm{h}}),\qquad L_{\max}=U/p.
\end{aligned}
\label{eq:screen-quantities}
\end{equation}
Requested corners also consume lap length and therefore restrict the feature combinations admitted by the screen. Let $\operatorname{MinArc}(f,r_{\min},w)$ denote the minimum arc length of feature $f$ under our construction rules. We require
\begin{equation}
\begin{aligned}
2\pi r_{\min} &\leq L\leq L_{\max},\\
\sum_{f\in\pi_{\mathrm{order}}}\operatorname{MinArc}(f,r_{\min},w)
&\leq \gamma(1-\rho)L.
\end{aligned}
\label{eq:screen}
\end{equation}
Here, $\rho=0.28$ reserves lap length for straights, while $\gamma=0.70$ limits the fraction of the remainder assigned to minimum feature arcs. Consequently, the arcs can occupy at most 50.4\% of the lap, leaving room for connectors, corner rounding, and loop closure. The feasibility screen combines this feature constraint with the estimated hall capacity and minimum turning radius.

For example, a request for a \SI{200}{\metre} lap in a $20\times15$\,m hall with $w=1.6$\,m, $t_{\mathrm{w}}=0.10$\,m, $d_{\mathrm{c}}=0.35$\,m, and $m_{\mathrm{h}}=0.5$\,m yields $L_{\max}\approx\SI{123.7}{\metre}$. We therefore reject this request before search because its target length exceeds the estimated capacity. However, the bounds constitute a conservative screen under our construction assumptions: rejection does not establish geometric impossibility, and admission does not guarantee that search will recover a valid map.
% Schema and feasibility defaults verified against tools/track_synth/prompt.py
% and spec.py in the author-supplied map-generator archive.
% Feature-classification details remain subject to the existing validation review.

\subsection{Coverage-Guided Track Construction}

We construct candidate tracks by covering the available grid while preserving a single closed corridor. Starting from the screened specification, \textsc{Discretise} determines the grid and corner radii, after which \textsc{SampleTree} samples a seeded tree over $2\times2$ cell blocks. \textsc{FormLoop} then joins local cycles along this tree, incorporates unused cells, and removes collinear vertices, yielding a closed waypoint loop.

Grid folding supplies candidate hairpins, whereas \textsc{AddFeatures} creates chicanes through paired opposite turns and sweepers by bevelling selected corners near $90^\circ$. To obtain a smooth centreline, \textsc{RoundCorners} replaces vertices with circular arcs and verifies that the adjoining tangent segments remain sufficiently long. We subsequently apply \textsc{CountFeatures} using the classification rules employed during final validation and resample candidates that violate clearance or feature-count constraints. Thus, coverage structures the search, while the final map-level checks determine whether the resulting geometry, including lap length, satisfies the requested tolerances.

\subsection{Raster-Level Validation}

Let $C:[0,L_C)\to\mathbb{R}^2$ be the rounded centreline and $d(\mathbf{x},C)=\min_s\|\mathbf{x}-C(s)\|_2$. Rasterisation assigns each square cell of width $\Delta$, centred at $\mathbf{x}$, the value
\begin{equation}
M(\mathbf{x})=
\begin{cases}
254, & d(\mathbf{x},C)\leq w/2-\Delta/2,\\
0, & w/2-\Delta/2<d(\mathbf{x},C)\leq w/2+t_{\mathrm{w}},\\
205, & \text{otherwise}.
\end{cases}
\label{eq:rasterisation}
\end{equation}
After \textsc{Rasterise} applies Eq.~\eqref{eq:rasterisation}, \textsc{CheckMap} evaluates the exported grid because rasterisation can alter corridor width and separation. An accepted map must contain exactly one corridor and one infield, match length within 3\% and width within 5\%, and satisfy the requested radius, clearance, feature counts and order, hall dimensions, and maximum straight length. Importantly, a failed specification screen and an exhausted search budget produce distinct failure reports. Figure~\ref{fig:req2map} summarizes this construction and validation loop.
% AUTHOR CONFIRMATION: length validation remains applied after coverage-guided generation.
% R4: numerical classifier/raster tolerances and feature-order enforcement remain open.

\begin{figure*}[t]
\centering
\resizebox{\textwidth}{!}{\input{figures/fig_track_construction.tex}}
\caption{Overview of the specification-guided construction pipeline. Feasible requests guide grid-based construction and smoothing, after which the exported occupancy map is validated before USD generation. Invalid maps return to structure sampling, whereas infeasible requests terminate before search.}
\label{fig:req2map}
\end{figure*}

\subsection{Automatic Environment Generation}
\label{sec:system}
\label{sec:problem}

We use a ROS occupancy map, together with its resolution and origin, as the common input to environment construction. Encoding occupied walls as 0, unknown background as 205, and free track as 254 allows recorded SLAM maps, rasterised circuits, and generated layouts to share one interface. The builder converts each accepted map into a USD stage that can be combined with our vehicle asset for simulation and learning, as summarized in Fig.~\ref{fig:pipeline}.

\label{sec:map2env}

Our builder extracts free space directly from labelled cells in filled maps and applies flood filling to exclude the exterior and infield in thin-line maps. It then smooths the corridor boundaries and sweeps them into tubes with a default diameter of \SI{33}{\centi\metre}, adding evenly spaced rings that reproduce RoboRacer air ducts. For collision geometry, the builder extracts a level set of the free-space distance field and extrudes its boundaries into invisible wall meshes. Crucially, the visual and collision representations remain separate, allowing appearance and contact geometry to be configured independently. The resulting meshes and floor are assembled into a metre-scale USD stage with configurable textures, colours, physical materials, and compliance parameters.

\label{sec:track-collection}

Using this common interface, we assemble a 130-track reference collection comprising 51 SLAM maps from competitions and testing sessions, 23 maps from the F1TENTH racetracks collection~\cite{f1tenth_racetracks}, 44 scaled Formula~1 circuits, and 12 layouts generated from explicit requirements.

For imported circuits, we first scale the supplied centrelines to the vehicle platform and subsequently smooth, resample, and rasterise them~\cite{tumftm_racetrackdb,bacinger_f1circuits}. Local narrowing prevents non-adjacent track sections from merging during rasterisation, thereby preserving the intended corridor topology.

\subsection{Vehicle and Learning Interface}
\label{sec:vehicle}

To connect simulation with physical deployment, we reconstruct a CAD model of the standard Traxxas-based RoboRacer from 3D scans of its components. The corresponding USD asset provides the vehicle used throughout our simulation environments (Fig.~\ref{fig:pipeline}). Table~\ref{tab:car21} summarizes its geometry, inertial parameters, and measured operating conditions.

We instantiate vehicle--track pairs programmatically in Isaac Sim~\cite{isaacsim}, using GPU PhysX with the TGS solver to model three-dimensional contact and suspension. GPU ray casting reproduces the physical lidar's \SI{270}{\degree} field of view and its alignment with the chassis. A ROS~2 bridge publishes odometry, transforms, and simulation time while accepting the physical car's Ackermann commands at configurable sensing and control rates.

The measurements in Table~\ref{tab:car21} characterize the vehicle over four competition runs on RoboRacer GP Track A, totaling \SI{62.4}{\minute}. These recordings are independent of the nine policy-deployment runs used in our evaluation. Dividing the lateral pull-test force by vehicle weight yields an effective tyre--surface friction estimate of $\mu\approx0.83$.

\begin{table}[t]
\caption{Physical RoboRacer parameters and operating statistics on GP Track A. CoM denotes centre of mass, $p_{99}$ the 99th percentile, and dashes unreported statistics.}
\label{tab:car21}
\tablestyle
\renewcommand{\arraystretch}{1.10}
\begin{tabular*}{\linewidth}{@{\extracolsep{\fill}}lrl@{}}
\toprule
\textbf{Parameter} & \textbf{Value} & \textbf{Unit} \\
\midrule
Wheelbase & 0.3275 & [m] \\
Track width & 0.265 & [m] \\
Mass & 4.885 & [kg] \\
Yaw inertia & 0.089 & [kg m$^2$] \\
CoM to front axle & 0.1753 & [m] \\
CoM to rear axle & 0.1522 & [m] \\
CoM height & 0.063 & [m] \\
Wheel radius & 0.0525 & [m] \\
Steering limit & 0.524 & [rad] \\
Tyre--surface friction $\mu$ & $\approx 0.83$ & [$-$] \\
\midrule
\multicolumn{3}{@{}l}{\textbf{Sensors and recording}} \\
Lidar beams & 1081 & [$-$] \\
Lidar field of view & 270 & [deg] \\
Scan rate & 39.9 & [Hz] \\
Odometry rate & 50.0 & [Hz] \\
Command rate & 20.0 & [Hz] \\
Recorded driving & 62.4 & [min] \\
\midrule
\end{tabular*}
\par
\begin{tabular*}{\linewidth}{@{\extracolsep{\fill}}lrrr@{}}
\textbf{Deployment metric} & \textbf{Median} & $\boldsymbol{p_{99}}$ & \textbf{Max} \\
\midrule
Speed [m/s] & 3.42 & 6.39 & 6.66 \\
Yaw rate [rad/s] & 1.20 & 5.46 & -- \\
\bottomrule
\end{tabular*}
\end{table}

\label{sec:policy}

Our experiments use residual reinforcement learning to maintain a common controller architecture across simulation and physical deployment. Following~\cite{trumpp2023rpl}, we add a learned correction to a Pure-Pursuit command:
\begin{equation}
\mathbf{a}_t=\mathbf{a}^{\mathrm{PP}}_t+\delta\mathbf{a}_t.
\label{eq:residual}
\end{equation}
The baseline follows the raceline, while the residual adjusts its command at each step. All deployed policies use the same residual-PPO design, clearance-shaped reward, three-frame observation, and nominal training-speed setting of \SI{1.5}{\metre\per\second}. Each frame contains 1080 lidar values, 11 vehicle-state values, and 60 look-ahead waypoint features, producing $3(1080+11+60)=3453$ inputs. Section~\ref{sec:policy-transfer} compares physical deployment of policies trained in F1TENTH Gym and RoboRacer Arena.

Both implementations apply a one-off collision penalty of $-50$ without rate scaling; however, their termination conditions reflect different simulation models. F1TENTH Gym uses a lidar-based inverse-time-to-collision heuristic, whereas RoboRacer Arena additionally applies a PhysX contact-force threshold of \SI{1.0}{\newton} and a \SI{5}{\milli\metre} beam margin. Both configurations set $\mu=0.8$, close to the pull-test estimate of $\mu\approx0.83$ in Table~\ref{tab:car21}. This coefficient enters Gym's single-track bicycle model and RoboRacer Arena's floor material, while the latter rigid-body model additionally represents 3D contact, suspension, and load transfer. Consequently, the comparison holds the policy architecture fixed but does not equate the underlying vehicle or collision models.
% AUTHOR REVIEW R1/R5: implementation details supplied by the author. Confirm
% exact detector logic, observation preprocessing,
% action bounds and speed-setting semantics. See editorial-review/consistency-review.md.

\section{Experimental Evaluation}
\label{sec:evaluation}

To evaluate the complete workflow, we investigate three questions: how reliably the pipeline satisfies track requirements, what computational cost construction and training require, and how the resulting policies behave on the physical car. We assess each stage separately and subsequently relate physical tracking error to achieved speed and lap completion. Map construction uses one AMD EPYC 9654 CPU core with peak memory of \SI{0.78}{\giga\byte}, while GPU measurements use an NVIDIA RTX PRO 6000 Blackwell Max-Q with \SI{96}{\giga\byte} VRAM.

\subsection{Specification Compliance}

We first evaluate whether the language models produce valid schemas while preserving the requested constraints. All four local models parse 12 requests covering units, feature order, and omitted fields. However, over 17 requests repeated three times, Mistral 7B introduces unstated values in six calls, whereas the remaining models preserve every constraint in all 51 calls (Table~\ref{tab:models}). We therefore select Gemma 4 31B, which achieves this observed accuracy with less memory than GPT-OSS 120B and lower latency than Qwen 3.6 35B.

\begin{table}[tb]
\caption{Constraint preservation across local language models. Parse comprises 12 requests, whereas Checks comprises 17 requests repeated three times. Runtime is reported per call with the model resident in memory.}
\label{tab:models}
\tablestyle
\begin{tabular*}{\linewidth}{@{\extracolsep{\fill}}lrrrr@{}}
\toprule
\raisebox{0.5em}{\textbf{Model}} & \raisebox{0.5em}{\textbf{Parse}} & \raisebox{0.5em}{\textbf{Checks}}
& \shortstack{\textbf{Time}\\\textbf{[s]}} & \shortstack{\textbf{VRAM}\\\textbf{[GB]}} \\
\midrule
Gemma 4 31B       & \textbf{12/12} & \textbf{51/51} & 9.3  & 19.0 \\
Qwen 3.6 35B      & \textbf{12/12} & \textbf{51/51} & 12.5 & 23.9 \\
GPT-OSS 120B      & \textbf{12/12} & \textbf{51/51} & 3.7  & 65.4 \\
Mistral 7B        & \textbf{12/12} & 45/51          & \textbf{0.8} & \textbf{4.4} \\
\bottomrule
\end{tabular*}
\end{table}

We next compare coverage-guided construction with differential evolution (DE)~\cite{storn1997de} and random sampling across ten specifications and three seeds per specification. A trial succeeds only when it returns a map accepted by the common raster validator. Coverage-guided construction succeeds in 70.0\% of the 30 trials, compared with 50.0\% for DE and 36.7\% for random sampling (Table~\ref{tab:constructors}). Moreover, it yields the lowest reported runtime, requiring \SI{3.7}{\second} per trial compared with \SI{5.1}{\second} and \SI{11.2}{\second}, respectively, including construction, map writing, and validation.

The comparison uses each procedure's native stopping rule: up to 800 candidates for coverage-guided search, convergence for DE, and one draw for random sampling. DE optimizes a curvature profile together with feature-turn and radius variables, enforcing total-turn and position closure while penalizing dimensional, clearance, hall-fit, and feature-count errors under a soft realism prior. Random sampling draws from the same admissible bounds without optimization or closure projection. Consequently, the experiment compares complete construction procedures under their respective budgets rather than equal numbers of candidate evaluations.

Because coverage-guided construction validates candidates before returning them, every emitted output satisfies the common validator. The remaining 30.0\% of trials exhaust the search budget without producing an accepted map. Thus, its success rate measures how often search reaches the acceptance condition, whereas a baseline may emit a map that subsequently fails the same checks.
% AUTHOR REVIEW R6: confirm where "built" is counted for each constructor.
% Algorithm 1 already validates coverage-guided candidates inside its retry loop.

\begin{table}[tb]
\caption{Comparison of map construction procedures over 30 trials per method. All outputs are evaluated by the same raster validator, while search budgets follow each procedure's native stopping rule. Runtime is reported in seconds per trial.}
\label{tab:constructors}
\tablestyle
\begin{tabular*}{\linewidth}{@{\extracolsep{\fill}}lrrrr@{}}
\toprule
\textbf{Constructor} & \textbf{Trials} & \textbf{Built} & \textbf{Valid}
& \textbf{Time} \\
\midrule
Coverage-guided   & 30 & \textbf{21} & \textbf{21} & \textbf{3.7} \\
DE optimiser      & 30 & 27          & 15          & 5.1 \\
Random sampling   & 30 & 30          & 11          & 11.2 \\
\bottomrule
\end{tabular*}
\end{table}

\begin{figure*}[!t]
\centering
\includegraphics[height=3cm,width=0.24\textwidth,keepaspectratio,trim=28bp 81bp 36bp 89bp,clip]{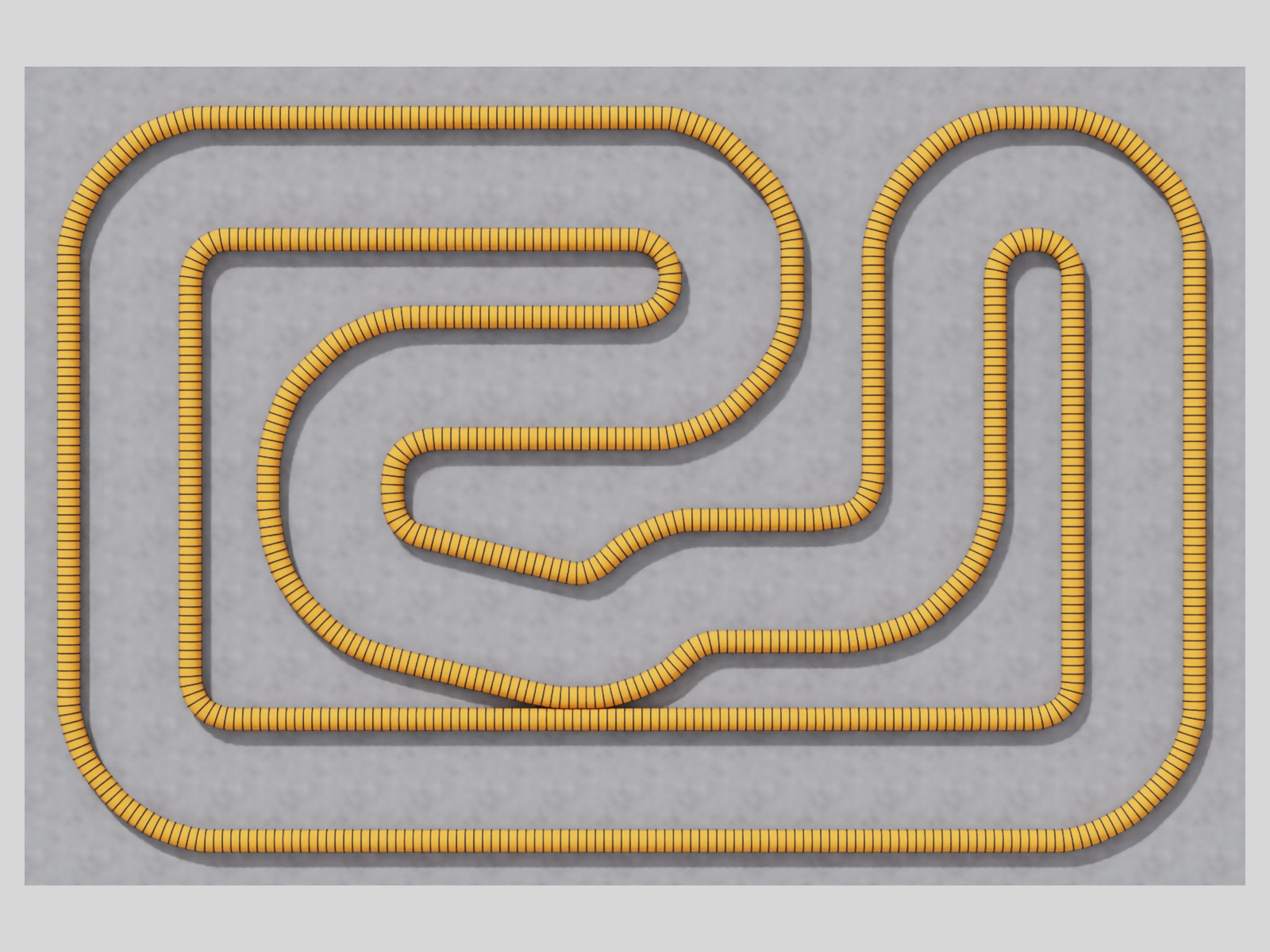}\hfill
\includegraphics[height=3cm,width=0.24\textwidth,keepaspectratio,trim=208bp 136bp 268bp 186bp,clip]{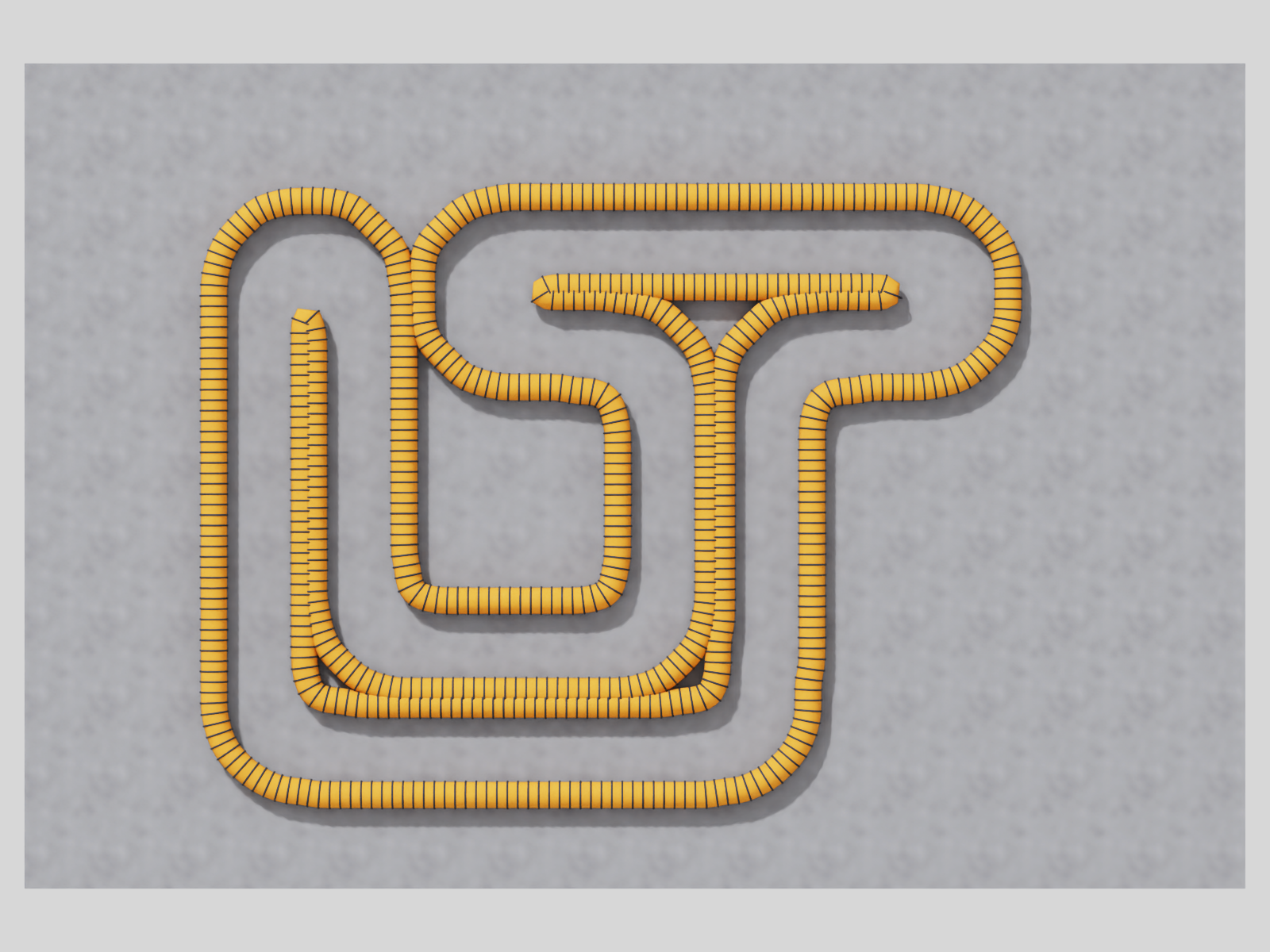}\hfill
\includegraphics[height=3cm,width=0.24\textwidth,keepaspectratio,trim=137bp 146bp 460bp 202bp,clip]{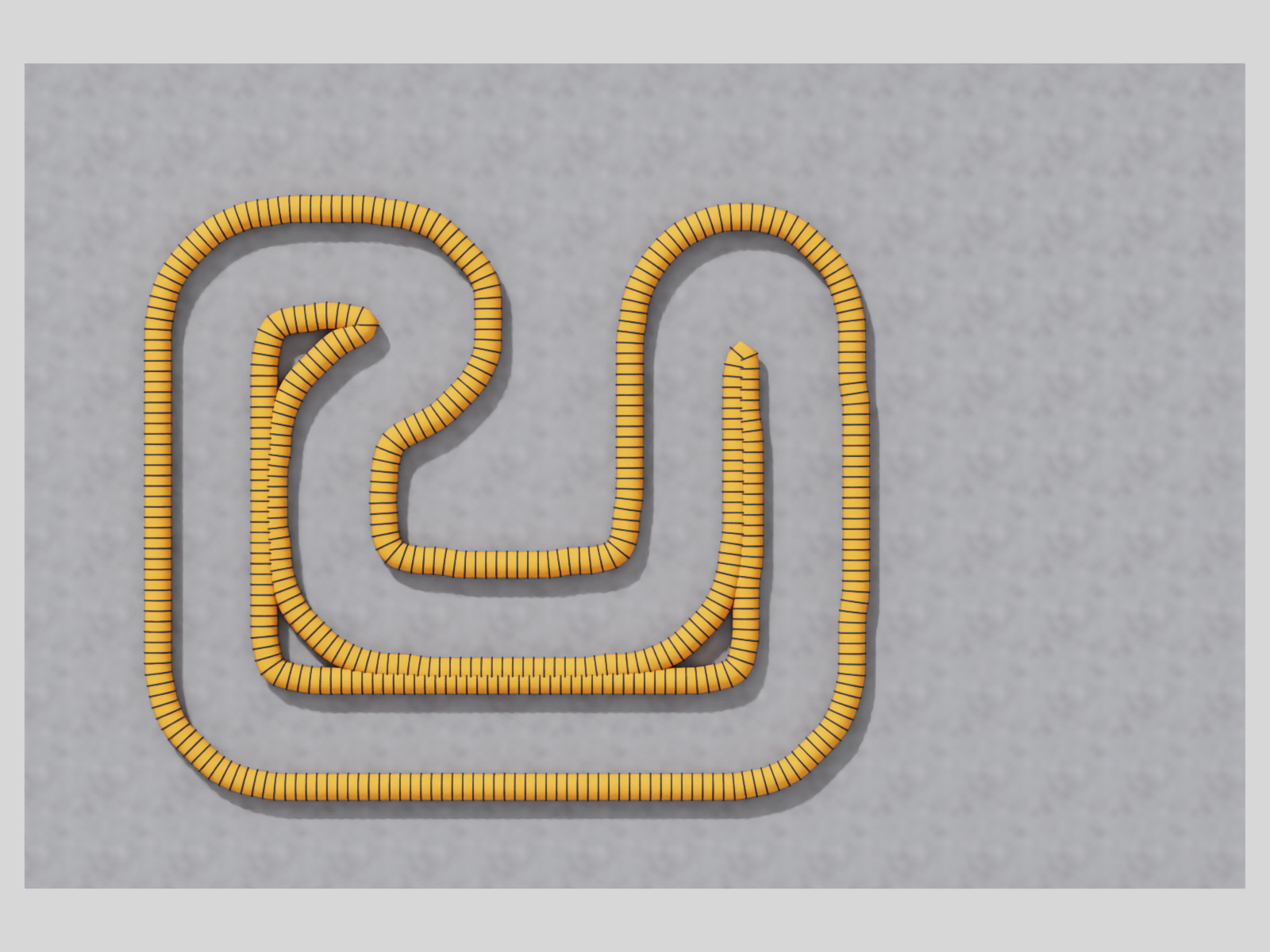}\hfill
\includegraphics[height=3cm,width=0.24\textwidth,keepaspectratio,trim=41bp 131bp 470bp 217bp,clip]{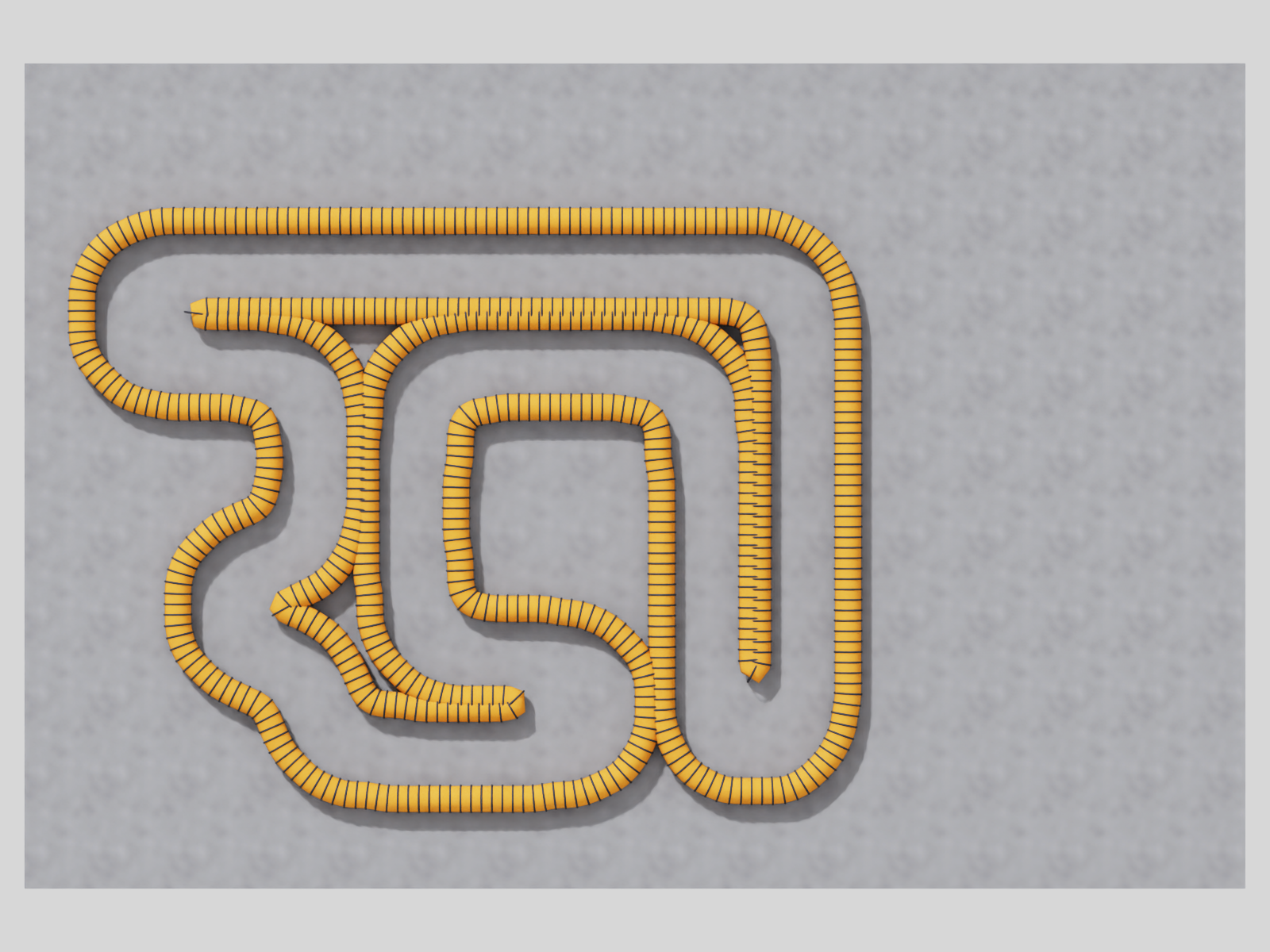}
\caption{Four layouts generated from the same request: a 45-m counter-clockwise lap of 1.6-m width, with three hairpins and one chicane in a $20\times15$-m hall.}
\label{fig:prompt-variants}
\end{figure*}

Finally, we evaluate the complete language-to-map pipeline on eight additional requests and obtain accepted maps for six. The feasibility screen rejects a \SI{200}{\metre} lap in a $20\times15$\,m hall, whereas a five-hairpin, \SI{45}{\metre} request passes the screen but exhausts the search budget. This distinction demonstrates why specification admission and successful construction must be reported separately. Figure~\ref{fig:prompt-variants} shows the geometric variation obtained from a shared specification.

\subsection{Conversion Cost and Geometric Fidelity}
\label{sec:circuits}

We measure conversion cost on eight maps spanning an 82-fold range in raster size. Constructing the corresponding USD environments requires \SIrange{1.18}{2.48}{\second} on one CPU core, representing only a 2.1-fold variation in runtime (Table~\ref{tab:map2usd}). Thus, asset construction remains a short preprocessing step for the tested inputs; however, the measurements exclude simulator startup and policy loading.

\begin{table}[tb]
\caption{Map-to-USD conversion across eight inputs. Length denotes centreline lap length at simulation scale, with imported circuits scaled to 1:10. Runtime uses one CPU core and excludes simulator startup.}
\label{tab:map2usd}
\tablestyle
\begin{tabular*}{\linewidth}{@{\extracolsep{\fill}}lrrrr@{}}
\toprule
\raisebox{0.5em}{\textbf{Map}} & \shortstack{\textbf{Length}\\\textbf{[m]}}
& \shortstack{\textbf{Raster}\\\textbf{[MP]}}
& \shortstack{\textbf{Build}\\\textbf{[s]}} & \shortstack{\textbf{USD}\\\textbf{[MB]}} \\
\midrule
RoboRacer GP         & 81.26  & 0.13  & 1.18 & 7.8 \\
Norisring            & 229.36 & 2.56  & 1.49 & 21.6 \\
Interlagos           & 430.40 & 3.02  & 1.68 & 40.5 \\
Monaco               & 332.86 & 3.04  & 1.60 & 31.3 \\
Zandvoort            & 431.58 & 3.48  & 1.75 & 40.6 \\
Silverstone          & 588.51 & 7.40  & 2.16 & 55.4 \\
Suzuka               & 580.05 & 8.43  & 2.20 & 54.4 \\
Spa-Francorchamps    & 699.84 & 10.77 & 2.48 & 65.9 \\
\bottomrule
\end{tabular*}
\end{table}

To assess geometric fidelity, we evaluate connectivity and centreline length across all 44 imported circuits. Every converted map retains a single connected free corridor, while centreline-length error averages 0.253\% and reaches at most 0.924\% relative to the supplied geometry. These results demonstrate that the tested conversion preserves both corridor topology and lap length.

\subsection{Parallel Training Throughput}
\label{sec:results}

To determine whether the constructed environments can supply sufficient experience for repeated learning experiments, we measure throughput during residual-policy training using the architecture in Sec.~\ref{sec:policy}. Steady-state logs include both experience collection and policy updates after warm-up. Table~\ref{tab:supply} reports these rates together with the track inputs supported by each platform, thereby connecting environment construction to the learning workload it enables.

\begin{table}[tb]
\caption{Comparison of training throughput and supported track inputs at the tested batch sizes. The official F1TENTH track collection is distributed separately~\cite{f1tenth_racetracks}.}
\label{tab:supply}
\tablestyle
\setlength{\tabcolsep}{3.5pt}
\begin{tabular*}{\linewidth}{@{\extracolsep{\fill}}lccc@{}}
\toprule
\raisebox{0.5em}{\textbf{Metric}} & \shortstack{\textbf{F1TENTH}\\\textbf{Gym}}
& \raisebox{0.5em}{\textbf{AutoDRIVE}} & \shortstack{\textbf{RoboRacer}\\\textbf{Arena}} \\
\midrule
Parallel environments & 24 & 20 & 256 \\
Aggregate steps/s & \num{1090} & 374 & \textbf{\num{7262}} \\
Steps per env. [Hz] & \textbf{45.4} & 18.7 & 28.4 \\
Track collection & 23 official & 6 predefined & 130 reference \\
Track generation & -- & -- & $\checkmark$ \\
Occupancy-map input & $\checkmark$ & -- & $\checkmark$ \\
Language input & -- & -- & $\checkmark$ \\
Map-to-3D build [s] & -- & -- & 1.2--2.5 \\
3D contact / rendering & -- & $\checkmark$ & $\checkmark$ \\
Gym interface & $\checkmark$ & -- & $\checkmark$ \\
\bottomrule
\end{tabular*}
\end{table}

RoboRacer Arena sustains \num{7262} aggregate steps/s across 256 environments, compared with \num{1090} across 24 environments for F1TENTH Gym and 374 across 20 environments for AutoDRIVE. The resulting aggregate rates are 6.7 and 19.4 times higher, while the respective per-environment rates are \SI{28.4}{\hertz}, \SI{45.4}{\hertz}, and \SI{18.7}{\hertz}. Thus, the tested RoboRacer Arena configuration supplies substantially more aggregate experience while retaining 3D simulation. Nevertheless, unequal batch sizes and software integrations prevent us from attributing this difference to the simulation backend alone.

In a separate benchmark, we examine the scaling limits of parallel contact physics using simplified rigid-body vehicles while excluding rendering and policy execution. Throughput peaks at \num{8707} vehicle-steps/s for 256 vehicles and subsequently decreases to \num{8368} at 1024 and \num{6818} at 2048. Additional replicas therefore produce diminishing returns in this workload. Together with the full-training measurements in Table~\ref{tab:supply}, these results establish the available simulation throughput; the following experiments evaluate the resulting policies on the physical vehicle.

\subsection{Physical Tracking and Lap Completion}
\label{sec:policy-transfer}

\begin{figure}[t]
\centering
\includegraphics[width=\linewidth]{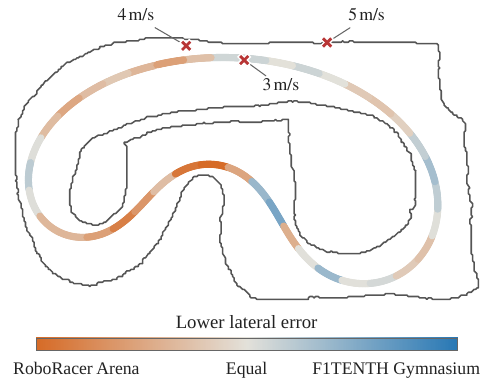}
\caption{Spatial distribution of tracking error on clean laps. Colour identifies the lower-error policy in each section, while crosses mark collisions of the Gym-trained policy. The RoboRacer Arena-trained policy completes all five runs without a crash.}
\label{fig:track-performance}
\end{figure}

\begin{figure*}[t]
\centering
\includegraphics[width=\textwidth]{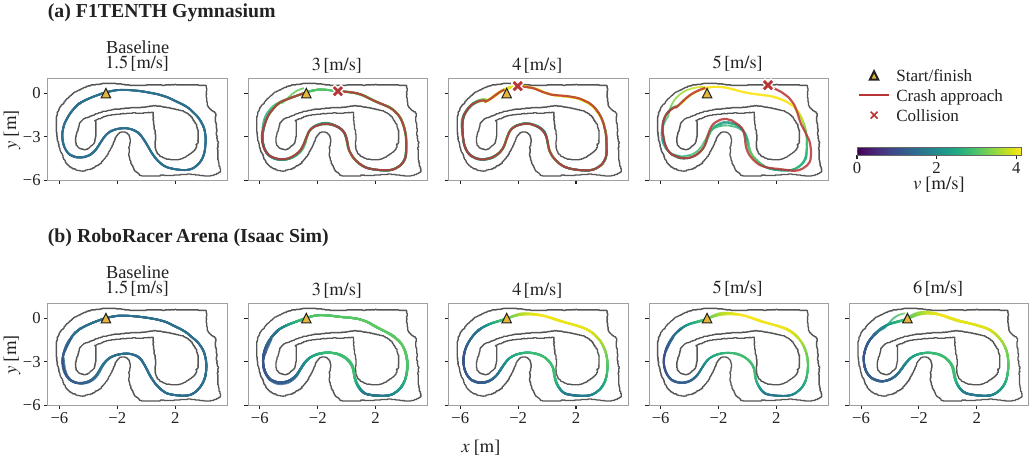}
\caption{Physical trajectories under increasing speed commands. The Gym-trained policy is shown above the RoboRacer Arena-trained policy. Columns include the 1.5-m/s baselines, colour denotes achieved speed capped at the pooled 99th percentile of clean-lap speed, and red marks collision approaches.}
\label{fig:per-take-comparison}
\end{figure*}

\begin{figure*}[t]
\centering
\includegraphics[width=\textwidth]{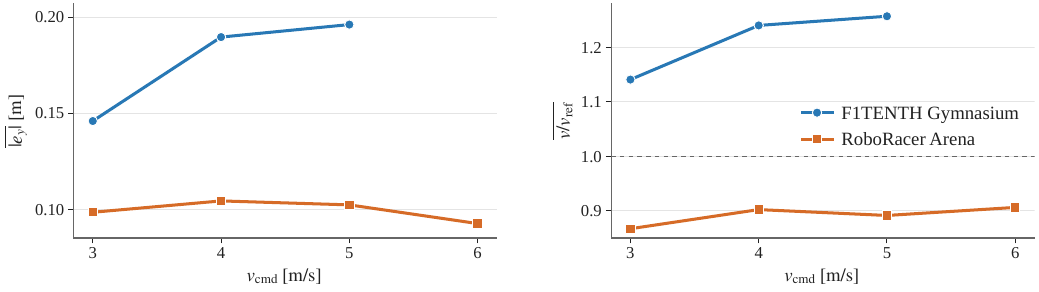}
\caption{Tracking and speed response across non-baseline commands. The RoboRacer Arena-trained policy maintains lower mean lateral error throughout the sweep. Speed ratio compares achieved speed with the local raceline target, and the dashed line denotes a ratio of one. Table~\ref{tab:policy-transfer} additionally includes the baseline runs.}
\label{fig:degradation-sweep}
\end{figure*}

Our final experiments investigate how tracking error, achieved speed, and lap completion change after deployment on the physical platform. We train residual-PPO implementations in F1TENTH Gym, RoboRacer Arena, and AutoDRIVE using five maps sampled with a fixed selection seed. Since the tested AutoDRIVE checkpoints complete no laps in simulation, we deploy only the remaining two policies. Nine ROS recordings, comprising four Gym-trained and five RoboRacer Arena-trained runs, provide localized pose, wheel-odometry speed, and steering commands. Figures~\ref{fig:track-performance}--\ref{fig:degradation-sweep} summarize the resulting errors, trajectories, and response to increasing speed commands.

We measure mean absolute lateral error relative to the raceline, $E=\overline{|e_y|}$, and mean speed ratio relative to the local target, $R=\overline{v/v_{\mathrm{ref}}}$. Table~\ref{tab:policy-transfer} aggregates these quantities with equal weight assigned to each run, including the baselines, whereas Fig.~\ref{fig:degradation-sweep} separates non-baseline runs by command. The reported minimum and maximum are extrema across run means and therefore describe between-run variation rather than instantaneous peaks.
% AUTHOR REVIEW R1: map names, selection seed, equality of simulator subsets,
% and inclusion/exclusion of the physical test track are still to be confirmed.

\begin{table}[tb]
\caption{Physical deployment results across all nine runs, including baselines. Statistics weight run means equally. $E$ denotes mean absolute lateral error, $R$ mean speed ratio, and $\sigma$ sample standard deviation; Min and Max are extrema of run means.}
\label{tab:policy-transfer}
\tablestyle
\setlength{\tabcolsep}{2.5pt}
% Reproduced from deployment-comparison-source/table_vi_statistics.csv.
\begin{tabular*}{\linewidth}{@{\extracolsep{\fill}}llrrrr@{}}
\toprule
\textbf{Training platform} & \textbf{Metric} & $\boldsymbol{\mu}$ & $\boldsymbol{\sigma}$ & \textbf{Min} & \textbf{Max} \\
\midrule
F1TENTH Gym & $E$ [m] & 0.148 & 0.062 & 0.062 & 0.196 \\
 & $R$ [$-$] & 1.063 & 0.304 & 0.612 & 1.257 \\
RoboRacer Arena & $E$ [m] & 0.093 & 0.014 & 0.069 & 0.104 \\
 & $R$ [$-$] & 0.832 & 0.134 & 0.595 & 0.906 \\
\bottomrule
\end{tabular*}
\end{table}

Across these runs, the RoboRacer Arena-trained policy achieves a \SI{37}{\percent} lower mean absolute lateral error, measuring \SI{0.093}{\metre} compared with \SI{0.148}{\metre} for the Gym-trained policy. Its error also varies less between runs, with $\sigma=\SI{0.014}{\metre}$ rather than \SI{0.062}{\metre}. However, the respective mean speed ratios of 0.83 and 1.06 show that the lower tracking error is accompanied by a more conservative speed response. Figure~\ref{fig:track-performance} localizes the corresponding errors and collisions along the track.

We next compare lap completion to determine whether each policy sustains its tracking behavior over repeated laps. The RoboRacer Arena-trained policy completes ten consecutive laps at each 4, 5, and \SI{6}{\metre\per\second} command, with no crashes across its five runs, including the \SI{1.5}{\metre\per\second} baseline and \SI{3}{\metre\per\second} run. In contrast, the Gym-trained policy completes its baseline but crashes after 9, 8, and 3 laps at 3, 4, and \SI{5}{\metre\per\second}, respectively (Fig.~\ref{fig:per-take-comparison}). The command settings do not represent sustained achieved speeds, and Gym's faster best clean lap, \SI{11.32}{\second} compared with \SI{12.16}{\second}, reveals a trade-off between peak lap pace and repeated completion.

The command sweep further shows how this trade-off evolves with requested speed. Gym's mean absolute lateral error rises from \SI{0.146}{\metre} to \SI{0.196}{\metre} over 3--5\,m/s, whereas RoboRacer Arena's error remains within \SIrange{0.093}{0.104}{\metre} over 3--6\,m/s with a speed ratio of 0.87--0.91 (Fig.~\ref{fig:degradation-sweep}). At the highest command of \SI{6}{\metre\per\second}, the policy therefore maintains stable tracking at four times the nominal training setting; however, achieved speeds remain below both the command and, on average, the local raceline target.

{\clubpenalty=10000
To examine behavior within a lap, we compare corner and straight samples pooled across clean laps, using the upper and lower quartiles of sampled absolute raceline curvature. The Gym-trained policy averages \SI{2.21}{\metre\per\second} in corners, 19\% above the local target, whereas the RoboRacer Arena-trained policy averages \SI{1.55}{\metre\per\second}, 17\% below it. The policies exceed the target in 74\% and 17\% of corner samples and reduce mean speed from straights to corners by 20\% and 41\%, respectively. Thus, RoboRacer Arena's stronger speed reduction coincides with lower lateral error in both regions: \SI{0.129}{\metre} in corners and \SI{0.060}{\metre} on straights, compared with Gym's \SI{0.201}{\metre} and \SI{0.092}{\metre}.\par}

RoboRacer Arena's more conservative speed response may arise from its termination conditions, its vehicle dynamics, or their interaction. Separate ablations are therefore required to identify their individual effects rather than attributing the observed transfer behavior to any single component.
% AUTHOR REVIEW R5: all supplied regional statistics reproduce with exact pooled
% clean-lap curvature quartiles. 2732 Gym and 5741 Isaac corner samples.
% These are sample-weighted quartiles, not unweighted reference-line quartiles.
% The forthcoming baseline/residual/final-command profile is needed to identify
% braking onset and distinguish learned braking from baseline or actuator effects.

\section{Discussion and Limitations}

RoboRacer Arena makes track design part of a repeatable learning experiment. A typed specification fixes the requirements, a seeded constructor proposes geometry, and the exported map determines acceptance. Since validation captures rasterisation-induced changes, the constraints remain tied to the representation supplied to the simulator. The conversion and throughput results demonstrate that these environments support parallel learning while geometry varies within a common vehicle and controller interface.

Physical deployment further demonstrates the practical value of the infrastructure. The RoboRacer Arena-trained policy sustains repeated laps under increased commands with lower tracking error, whereas the Gym-trained policy achieves a faster best clean lap. Together, these results show that the workflow can produce policies that remain stable under demanding physical tests. Nevertheless, matched-speed performance and generalization across tracks and vehicles remain unresolved.

Building on these promising results, future work will study track geometry, vehicle modelling, and controller design under matched budgets and simulator workloads. Isolated changes to friction, motor models, and termination rules may explain the observed transfer behavior, while learned primitives could expand geometric diversity. Extending the planar representation with Gaussian-splat reconstruction and collision geometry may further support vision-based learning. The planned release will provide the tools, vehicle models, and tracks required for these studies.

\section{Conclusion}

In this paper, we presented RoboRacer Arena, a specification-driven framework that connects explicit requirements to validated occupancy maps and autonomous-racing simulation assets. Coverage-guided construction, a common map-to-USD builder, and a CAD-modelled vehicle support 130 reference tracks and parallel policy training. RoboRacer Arena achieves the strongest valid-map generation result among the tested procedures under their respective budgets, while physical deployment yields ten consecutive laps at commands up to four times the nominal training setting. Although the resulting policy exhibits a conservative speed response, these results demonstrate the practical potential of treating track geometry as a controlled experimental variable. Building on this foundation, the planned release will enable reproducible studies of how geometry and simulation choices affect learned racing behavior.
\section*{ACKNOWLEDGMENT}
This research was funded in whole or in part by the Austrian Science Fund (FWF)
10.55776/DOC1345324.
% \section*{Acknowledgment}

% Claude Opus 5 assisted with software subroutines, while ChatGPT-6 Astra assisted with manuscript editing.

\bibliographystyle{IEEEtran}
\bibliography{refs}
\end{document}